\documentclass[letterpaper, 10 pt, journal, twoside]{IEEEtran}
\usepackage{cite}

\usepackage{amsmath,amsfonts}

\usepackage{array}

\usepackage{stfloats}

\usepackage{url}

\usepackage[numbers]{natbib}
\usepackage{caption}
\usepackage{subcaption}
\usepackage{multirow}

\usepackage{algorithm}
\usepackage[noend]{algpseudocode}
\usepackage{array}
\usepackage{booktabs}
\usepackage{graphicx}
\usepackage[colorlinks=true, urlcolor=black, citecolor=green]{hyperref} 
\usepackage{multirow}

\usepackage{textcomp}
\usepackage{verbatim}
\usepackage{xfrac}

\usepackage{xstring,xifthen}
\newcommand{\subsubsec}[1]{\vspace{0.2em}\noindent\textbf{\IfEndWith{#1}{.}{#1}{#1.}}}

\usepackage{xspace}
\makeatletter
\DeclareRobustCommand\onedot{\futurelet\@let@token\@onedot}
\def\@onedot{\ifx\@let@token.\else.\null\fi\xspace}
\def\eg{\emph{e.g}\onedot} 
\def\ie{\emph{i.e}\onedot}

\makeatother
\DeclareMathAlphabet\mathbfcal{OMS}{cmsy}{b}{n}

\begin{document}

\title{On the challenges to learn from \\ Natural Data Streams}
\author{ 
Guido Borghi, Gabriele Graffieti and Davide Maltoni
\thanks{
The authors are with the Department of Computer Science and
Engineering, University of Bologna, 40126 Bologna, Italy (e-mail:
guido.borghi@unibo.it; gabriele.graffieti@unibo.it; davide.maltoni@unibo.it).
}

}

\maketitle

\begin{abstract}
In real-world contexts, sometimes data are available in form of Natural Data Streams, \ie data characterized by a streaming nature, unbalanced distribution, data drift over a long time frame and strong correlation of samples in short time ranges. Moreover, a clear separation between the traditional training and deployment phases is usually lacking.
This data organization and fruition represents an interesting and challenging scenario for both traditional Machine and Deep Learning algorithms and incremental learning agents, \ie agents that have the ability to incrementally improve their knowledge through the past experience. 
In this paper, we investigate the classification performance of a variety of algorithms that belong to various research field, \ie Continual, Streaming and Online Learning, that receives as training input Natural Data Streams.
The experimental validation is carried out on three different datasets, expressly organized to replicate this challenging setting.  
\end{abstract}


\IEEEpeerreviewmaketitle

\section{Introduction}\label{sec:introduction}
The ability to incrementally learn from \textit{Natural Data Streams} is a particularly desirable skill for those AI agents, \ie systems able to learn from the experience, that operates in real-world contexts~\cite{parisi2019continual, graffieti2022continual}. 
Natural Data Streams are identified as continuous data sequences acquired by these learning agents, in which temporally close samples can be very similar and, at the same time, the same object or scene can abruptly vary in long-time acquisitions.
This is the case, for instance, of a domestic robot that acquires the surrounding environment and classifies everyday objects, that may change their visual appearance due to different light sources (indoor or outdoor). Another example can be a vision-based system for self-driving cars that classifies vehicles: for instance, the data may vary depending on the road type, the weather, the part of the day.

To maintain or even improve a reliable classification capability, these learning agents should be able to quickly adapt their knowledge to the surrounding world, which continuously evolves and visually change.
In this context, the learning process greatly differs from the classic Machine Learning setting, in which the agent's life is usually divided into two sequential, fixed, and separated phases: \textit{training} and \textit{deployment}.
In a real-world context, these two phases are not realistic since all training data are not available at the beginning of the learning phase, and there is not a single deployment procedure. On the contrary, the agent must be able to continue to learn from new observations, updating its knowledge continually, without a definite boundary between training and deployment.

Indeed, from a formal point of view, the learning paradigm based on Natural Data Streams is based on the following peculiar elements:
\begin{enumerate}
    
    \item \textbf{Stream}: input data is available as a continuous temporal stream.
    Formally, in a certain time instant $t$, only a single data point $x_t$ is available as input. Then, given a time range $[t, t+k]$, multiple sequential data samples are available for input $\{x_t, x_t+1, …,  x_t+k\}$.
    The streaming nature of training data makes it harder to have a pre-defined training set, which fully represents the current and future distribution of exploited data. The learning phase must therefore be continuous and distributed over time~\cite{hayes2020lifelong}.
    
    \item \textbf{Natural}: the distribution of the input data is influenced by the realistic fruition and interaction with the environment by the learning agent. Therefore, it is easy to foresee that the data can be characterized by a high visual similarity in a short time range (\eg the domestic robot is handling a particular object), which can abruptly change in long-time ranges (\eg the light source has changed)~\cite{graffieti2022continual}. The natural data acquisition influences also the class distribution: it is likely that objects are not uniformly available in input data. These important aspects are further discussed and analyzed in Section~\ref{sec:main_features}.
    
\end{enumerate}

Finally, it is desirable that these learning agents, especially if implemented in embedded systems, \ie systems with some hardware limitations, such as computational power, memory size, power consumption and similar, have also the following characteristics, commonly assumed in the literature~\cite{hayes2020lifelong, hayes2019memory}:
\begin{itemize}
    
    \item \textbf{Fast update}: new knowledge, \ie new available input data $\{x_t, x_t+1, …,  x_t+k\}$, should be used immediately. 
    For this reason, it is advisable to accumulate only a relatively small amount of input data samples before triggering the learning process of the agent. We observe that the definition of the proper amount of data is tricky, since the computational load, the training time and the required memory are directly related.
    
    \item \textbf{Single-epoch learning}: a learning agent sees every data point $x_t$ only once. The learning procedure should be fast, then not based on multi-epoch learning iterations, a common setting in the Machine and Deep Learning scenarios.
    
    \item \textbf{Limited replay}: a learning agent must limit the memory usage, in terms of the number of input samples stored in the replay memory, even though several studies~\cite{parisi2019continual,graffieti2022continual} have shown that the replay memory efficiently contrasts the catastrophic forgetting~\cite{mccloskey1989catastrophic}.
    
\end{itemize}

In this paper, we aim to investigate the impact of Natural data Streams on several state-of-the-art methods available in the literature that, belonging to similar research topics, \ie Streaming Learning~\cite{hoens2012learning}, Continual Learning~\cite{parisi2019continual} and Online Learning~\cite{perry2011online}, have been created to handle a continuous learning procedure and rely on streaming input data.
We perform experimental evaluations exploiting three existing datasets, namely Soda10M~\cite{han2021soda10m}, Core50~\cite{lomonaco2017core50} and OpenLoris~\cite{she2019openlorisobject}, adapted to produce unbalanced streaming input data and temporal similarity or dissimilarity, accordingly to the above definition of Natural Data Stream.
For the sake of comparability, we release the source code and the complete dataset configurations. 
To the best of our knowledge, this is one of the first works that investigates on Natural Data Streams, focusing on temporal similarity and unbalanced data distribution, and we believe that this preliminary investigation can be useful to formally define the problems and detect the challenges that should be addressed in the future of this research field.



\section{Proposed Benchmarks}


Following \cite{graffieti2022continual}, we define two concepts that we are going to use in the rest of the paper:
\begin{itemize}
    
    \item \textbf{Experience}: an amount of data that the learning agent can accumulate in its temporary buffer before triggering the model update, \ie starting the learning procedure. 
    In all defined benchmarks, data belonging to different experiences are not shared, and, in particular, data in past experiences are not available in the future, with the exception for those data stored in the replay memory.
    
    \item \textbf{Macro-experience}: a set of single experiences acquired in similar conditions, therefore data from the same macro-experience presents similar characteristics, while data from different macro-experiences may greatly differ from each other. In our benchmarks, it is important to note that the learning agent is agnostic with respect to the boundaries across different macro-experiences. In other words, macro-experiences are a virtual division of input data, introduced only for testing purposes (\ie testing the model in specific moments during the training process to evaluate the performance). 
    
\end{itemize}

\subsection{Natural Data Stream Organization} \label{sec:main_features}
Two aspects are directly related to the nature of Natural Data Streams: temporal similarity and unbalanced data distribution. These aspects precisely affect the composition and the organizatio of the input streaming data as follows:

\begin{itemize}
    \item \textbf{Temporal Similarity}: due to the streaming nature of data, input samples can be highly temporally correlated in short time ranges. 
    From a formal point of view:
    \begin{equation}
        \lvert \text{fe}(x_t)-\text{fe}(x_{t+k}) \rvert \propto k
    \end{equation}
    where $\text{fe}$ is a feature extractor (for instance, a network trained on the Imagenet~\cite{deng2009imagenet} dataset) and $x_t$ is a particular data sample available in the range $[t,...,t+k, ...t+n]$.
    In other words, short-range consecutive input samples may be characterized by very similar visual appearance while data drifts can happen in long-time ranges.
    Therefore, a sequence of learning experiences, \textit{i.e.} a training procedure conducted on a certain amount of input data, can be aggregated in a single macro-experience as defined above. 
    In case of data drift, a switch from one macro-experience to another one happens. In this case the conditions can completely change but the class knowledge needs to be properly adjusted without forgetting, since old conditions are likely to return in the future.
    
    \item \textbf{Unbalanced Data Distribution}: the input sequence is likely to contain non-iid, or unbalanced, data, \ie for each class a different amount of data is available in a specific time range. It is important to observe that unbalanced data can characterize a single experience, but also a macro-experience. 
    In other words, it is not guaranteed the learning agent receive in input a balanced amount of data belonging to each class, and this is a challenging element for a proper and effective learning procedure~\cite{batista2004study,japkowicz2000learning}.

\end{itemize}

\begin{figure*}
    \centering
    \includegraphics[width=1\linewidth]{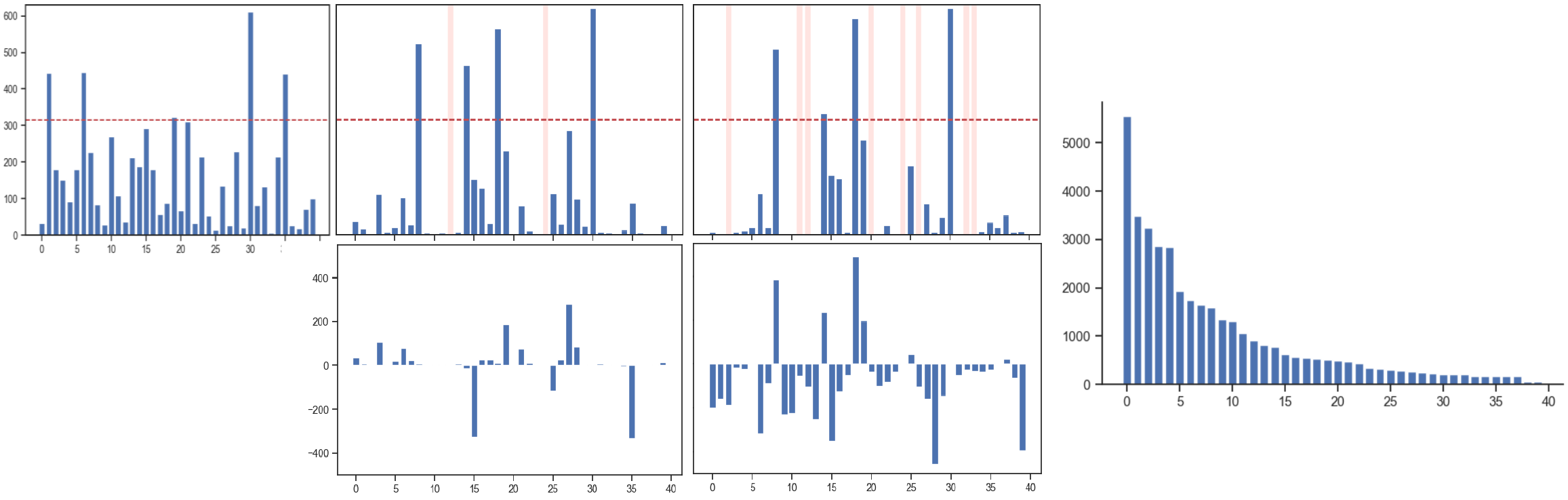}
    \caption{Data distributions of our version of the OpenLoris~\cite{she2019openlorisobject} dataset, adapted to represent a Natural Data Stream. On the left, the distribution of three different macro-experiences is reported (top row), including the variance for each class (second row), computed as the difference of the amount of the instances w.r.t. the previous macro-experience. Red columns denote the specific class is not available. On the right, the overall class distribution on the whole dataset, inspired by the Zipf's law~\cite{li2002zipf}.}
    \label{fig:openloris}
\end{figure*}

\subsection{Datasets}
To replicate the use of Natural Data Streams as input data, we were forced to select only datasets with temporal sequences of data, \ie images are not randomly sampled or acquired in different contexts, as in the majority of datasets (\eg ImageNet~\cite{deng2009imagenet}, MNIST~\cite{lecun1998mnist}, Cifar-10/100~\cite{krizhevsky2009learning}) commonly used in Continual Learning, but are organized in temporally ordered video frames. 
In the following, we detail the dataset selection in addition to the procedure that we eventually apply to adapt the dataset on the Natural Data Streams setting.

\begin{itemize}
    
    \item \textbf{OpenLoris}~\cite{she2019openlorisobject}: this robotic vision dataset consists of several videos collected with an RGB-D camera mounted on different mobile robots moving in real environments. This dataset presents realistic challenges that usually a robot faced, such as illumination changes (the illumination can significantly vary across time), the presence of occlusions (a part of the object to be classified can be occluded by other objects), variations in object size and camera-object distance, and cluttering (other objects that can interfere with the classification task are close to the main object). These challenges are quantized at different levels. $40$ different classes are available, and for each instance, up to $25$ seconds video (at $30$ fps) has been recorded with the depth camera for a total of around $500$ to $750$ frames. In our experiments, we consider RGB frames, maintaining the original training and testing splits. 
    
    
    We create an unbalanced version of this dataset exploiting the \textit{Zipf}'s law~\cite{li2002zipf}. 
    Indeed, as suggested in recent literature~\cite{chan2022zipfian}, the \textit{Zipf}'s law is a good approximation of how we naturally interact with objects in the environment. More in detail, we interact for the majority of the time with few common objects, while we interact sporadically with all the others. 
    We apply the Zipf's law on the whole dataset, obtaining a data distribution as depicted in the right part of Figure~\ref{fig:openloris}. From this general distribution of data across the whole dataset, we produce $9$ different macro-experiences. As detailed in the left part of Figure~\ref{fig:openloris}, the data distribution of each macro-experience is different from each other, and some classes may be not present in some macro-experiences.
    
    \item \textbf{Core50 UB}~\cite{lomonaco2017core50}: this dataset collects videos of hand held items acquired in $11$ sessions ($8$ indoor and $3$ outdoor) with different background types. A total of $50$ objects that belong to $10$ different categories are included. To increase the level of complexity, in our experiments we consider categories as classes: in this way, in the same class, there are objects even a lot different from a visual point of view (\eg different models and types of light bulbs belong to the same category). In addition, objects are partially occluded by the grasping hand. We split the paper as reported in the original paper, \ie we put $8$ sessions for training and $3$ sessions for testing In this setting, each session corresponds to a different macro-experience. 
    
    We obtain a Natural Data Stream implementing the procedure adopted in~\cite{graffieti2022continual}: in each macro-experience, not all classes are always available; in addition, only a certain proportion of the original samples for each class is available. 
    The input stream consists of video sequences that can be interleaved at frame level with other sequences: then, the temporal coherence is quite coarse, since in a given time range only frames belonging to the same object are temporally connected. 
    We observe that in this case, differently from the OpenLoris dataset, the same data distribution is used for each macro-experience. 
    
    \item \textbf{Soda10M}~\cite{han2021soda10m}: is a large-scale dataset acquired in the automotive setting. Indeed, there are $6$ classes (pedestrian, bicycle, car, truck, tram, and tricycle) acquired across $32$ Asiatic cities. 
    Object of interest are in provided bounding boxes manually annotated, for a total of $22$ object-centered.
    In this dataset, different macro-experiences correspond to scenes acquired in different hours (\textit{e.g.} day/night), weather conditions (\textit{e.g.} sun, rain) and different road types (\textit{e.g.} highway, country roads).
    The main challenges of this dataset are related to occlusions (images are acquired through a camera located in the frontal part of the car, in a low position), and the variability of the distance between objects and the camera that can negatively affect the spatial resolution of objects to be classified.
    
    The Natural Data Stream version of this dataset derives from the SSLAD competition~\cite{sslad}, in which the class ``car'' represent the large majority of the data distribution, while the class ``tricycle'' is largely underrepresented.
    The input stream consists of the sequence of the bounding boxes localized in images: since a single frame can have one or more bounding boxes, the resulting streaming has a huge variability in terms of the number of classes available in a certain time range.
    We observe that this stream setting is quite different from the previous ones, and then represents an interesting and challenging final test for the investigated algorithms.
    
\end{itemize}

For OpenLoris and Core50 datasets we create also a balanced version, \textit{i.e.} we maintain the same amount of data of the unbalanced versions but we balance the number and the composition of each class. In this manner, it is possible to train a method with the same amount of training data, varying only the data distribution. 

\subsection{Experimental Setup}
We organize the experiments in two different steps, in order to investigate the impact of Temporal Similarity and Unbalanced Data Distribution that characterize Natural Data Streams (see Section~\ref{sec:main_features}) on state-of-the-art algorithms.
Specifically, the steps are detailed as follows:

\begin{itemize}

    \item \textbf{Temporal Similarity vs Shuffling}: in this experiment, we aim to assess the effects of temporal similarity on input streaming data. In particular, we compare methods running the learning phase on the balanced versions of Core50 and OpenLoris datasets with and without shuffling data before training. In that way, we aim to highlight and isolate the impact of temporal similarity on the final performance of classifiers.
    
    \item \textbf{Balanced vs Unbalanced}: once investigated the impact of temporal similarity, the second experiment is designed to assess the impact of the unbalanced distribution of input data. 
    In this way, algorithms are tested on data that are close to Natural Data Streams, since input stream includes both temporal similarity and unbalanced data distribution.
    Then, we compare methods running the learning phase on data affected by temporal similarity (\textit{i.e.} data are not shuffled), on the balanced and unbalanced versions of Core50 and OpenLoris datasets. 
    As described, it is important to note that these balanced versions of each dataset consist of the same amount of data of the unbalanced ones. 
    
\end{itemize}

\section{Experiments}

\begin{figure*}[th!]
    \centering
    \begin{subfigure}[b]{0.24\linewidth}
        \centering
        \includegraphics[width=\linewidth]{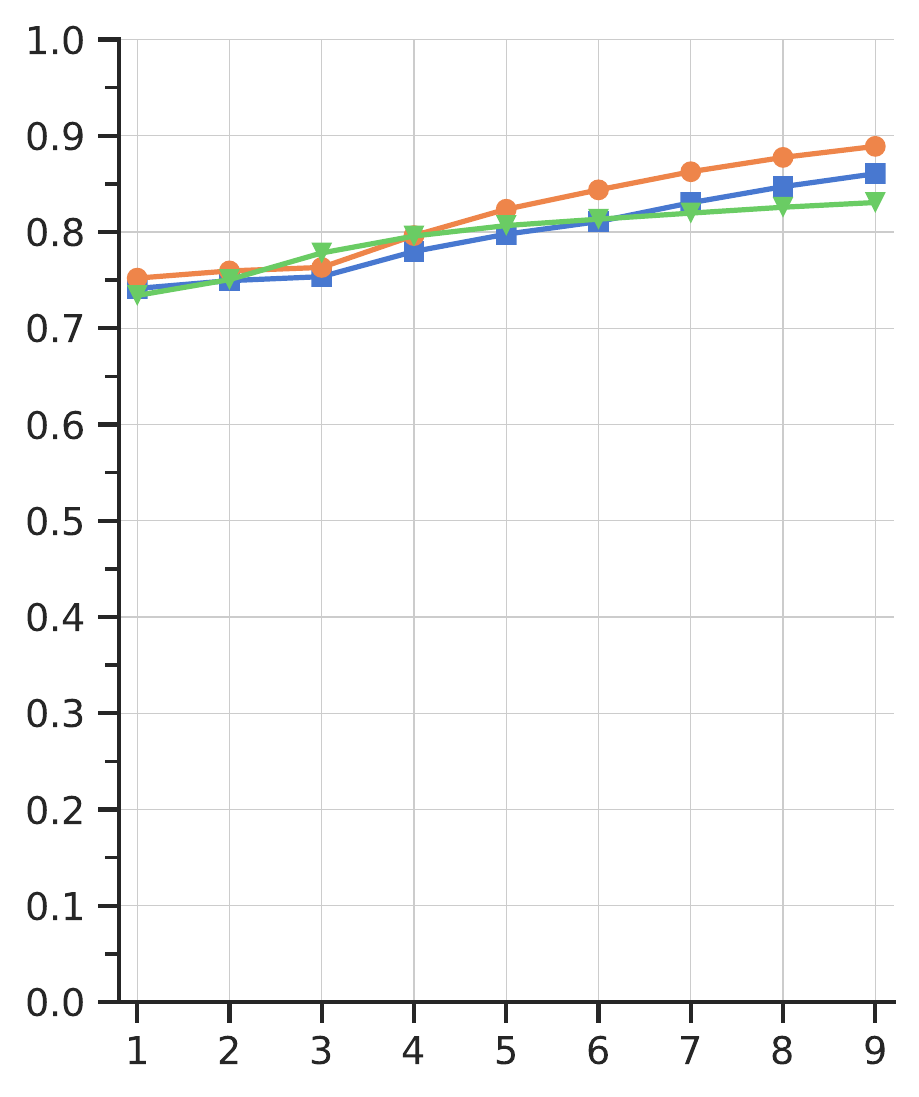}
        \caption{SLDA~\cite{hayes2020lifelong}}
        \label{fig:fd-dlib}
    \end{subfigure}
    \hfill  
    \centering
    \begin{subfigure}[b]{0.24\linewidth}
        \centering
        \includegraphics[width=\linewidth]{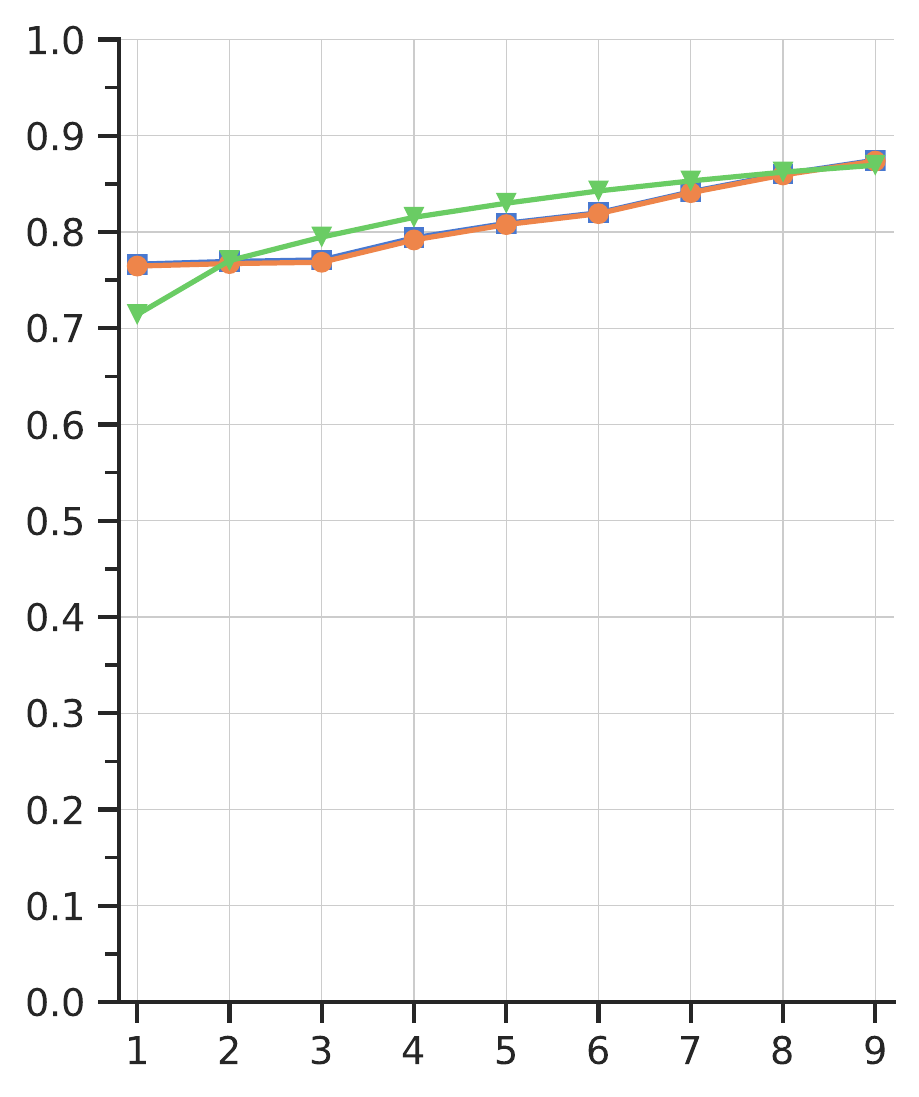}
        \caption{ExStream~\cite{hayes2019memory}}
        \label{fig:fd-dlib}
    \end{subfigure}
    \hfill   
        \centering
    \begin{subfigure}[b]{0.24\linewidth}
        \centering
        \includegraphics[width=\linewidth]{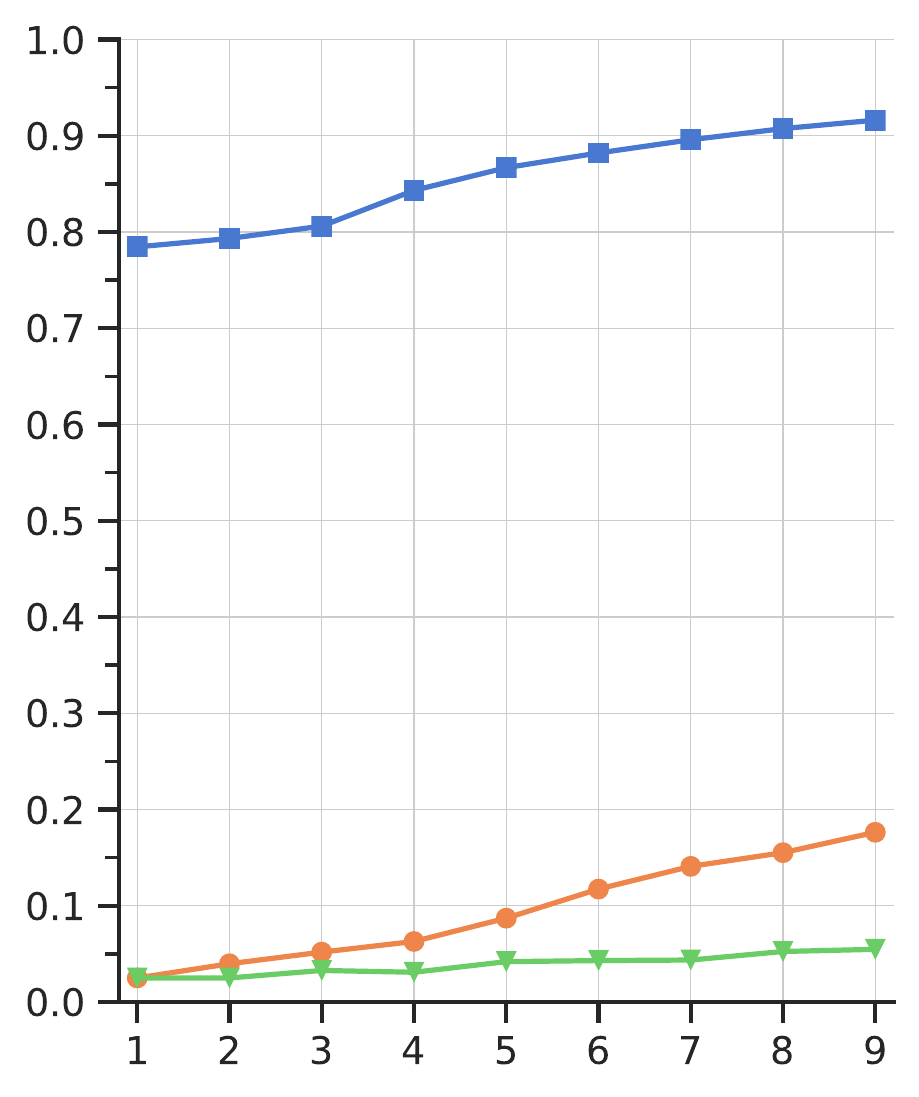}
        \caption{LwF~\cite{li2017learning}}
        \label{fig:fd-dlib}
    \end{subfigure}
    \hfill
        \centering
    \begin{subfigure}[b]{0.24\linewidth}
        \centering
        \includegraphics[width=\linewidth]{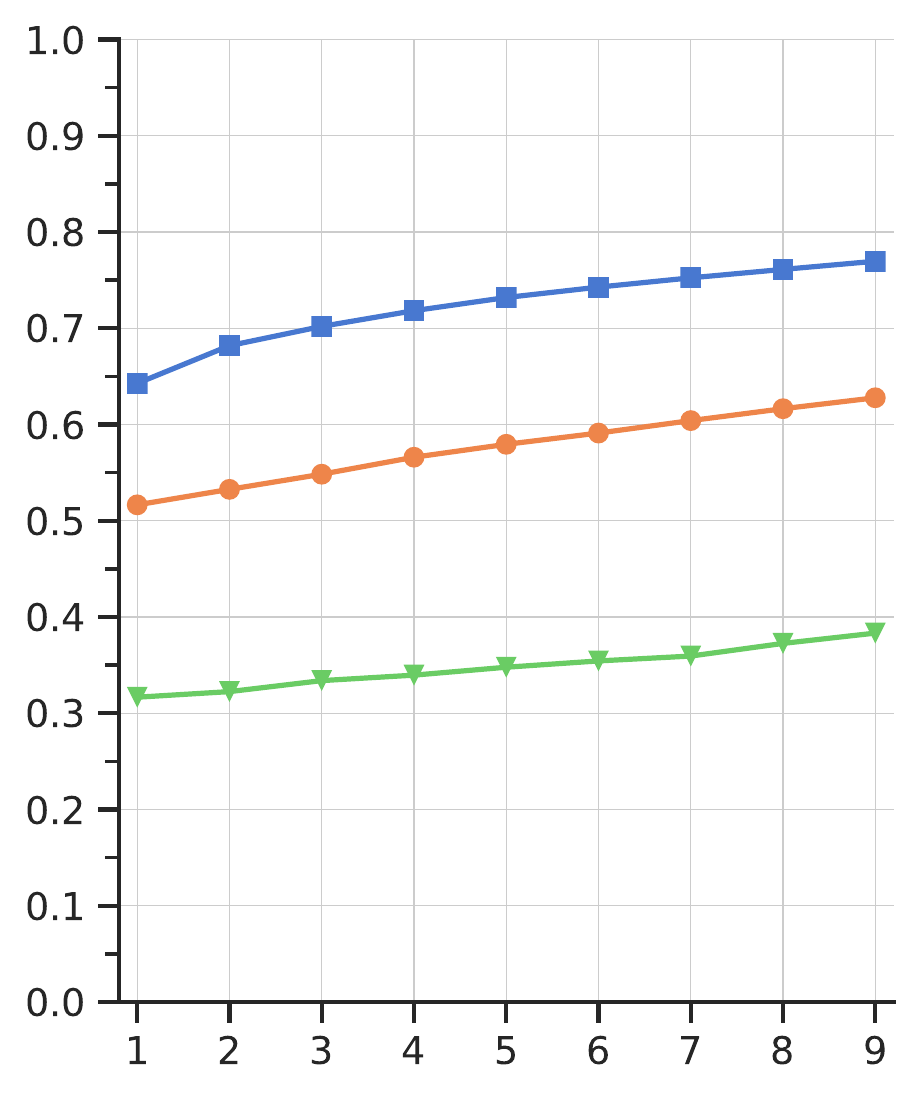}
        \caption{SSLAD~\cite{graffieti2022continual}}
        \label{fig:fd-dlib}
    \end{subfigure}
    \hfill
        \centering
    \begin{subfigure}[b]{0.24\linewidth}
        \centering
        \includegraphics[width=\linewidth]{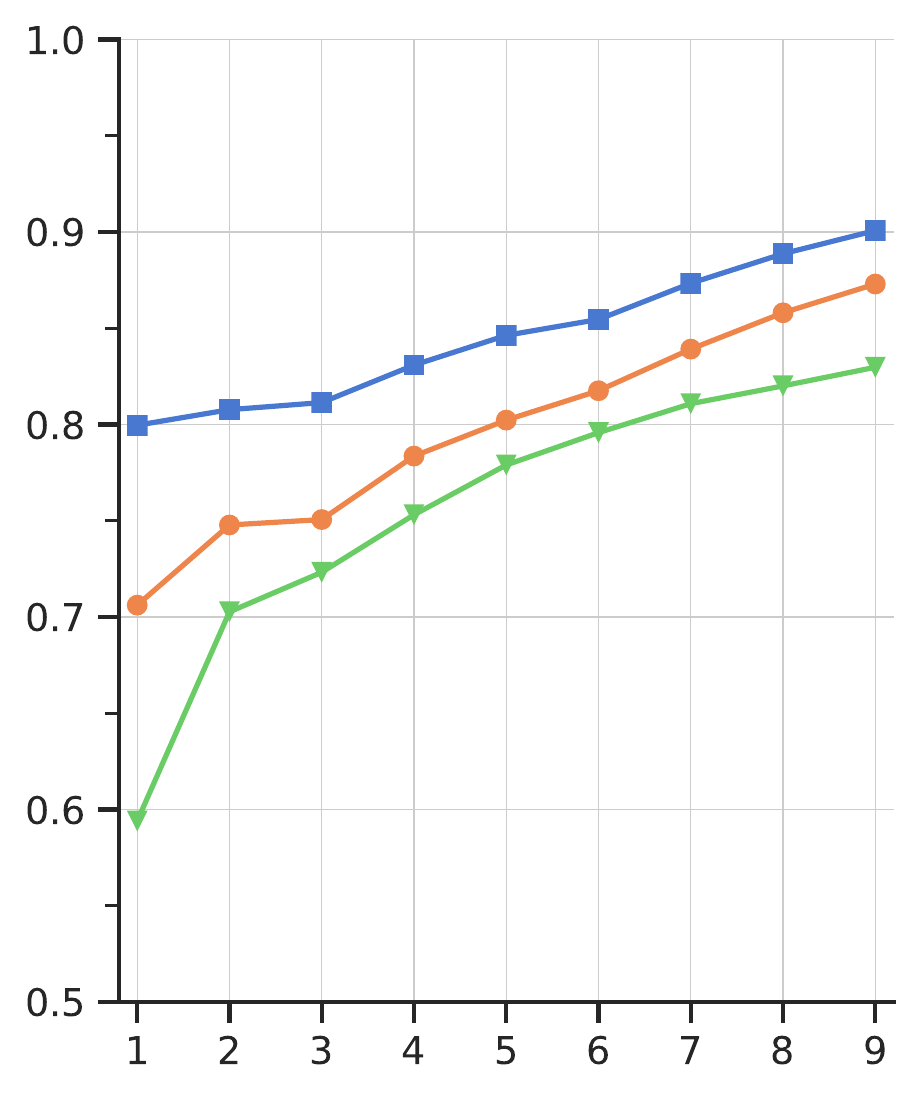}
        \caption{CBRS~\cite{chrysakis2020online}}
        \label{fig:fd-dlib}
    \end{subfigure}
    \hfill
        \centering
    \begin{subfigure}[b]{0.24\linewidth}
        \centering
        \includegraphics[width=\linewidth]{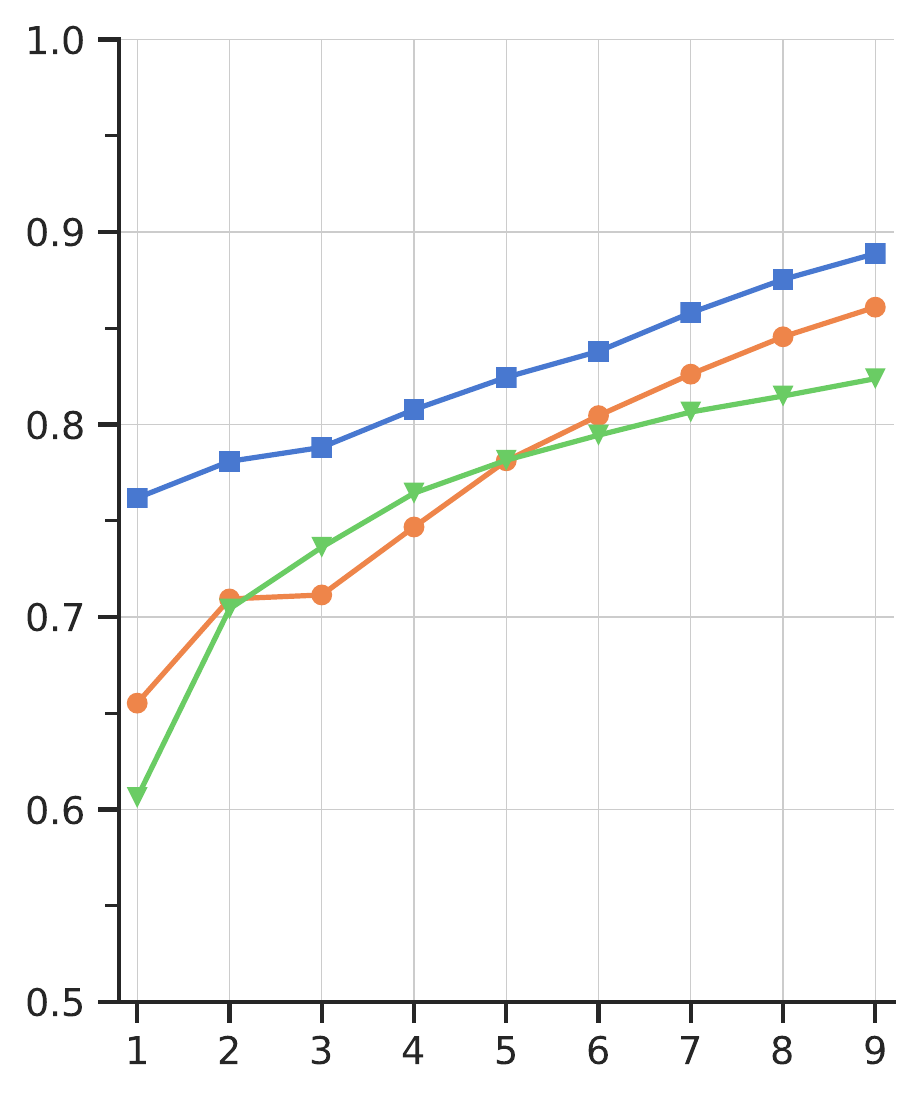}
        \caption{GSS~\cite{aljundi2019gradient}}
        \label{fig:fd-dlib}
    \end{subfigure}
    \hfill
        \centering
    \begin{subfigure}[b]{0.24\linewidth}
        \centering
        \includegraphics[width=\linewidth]{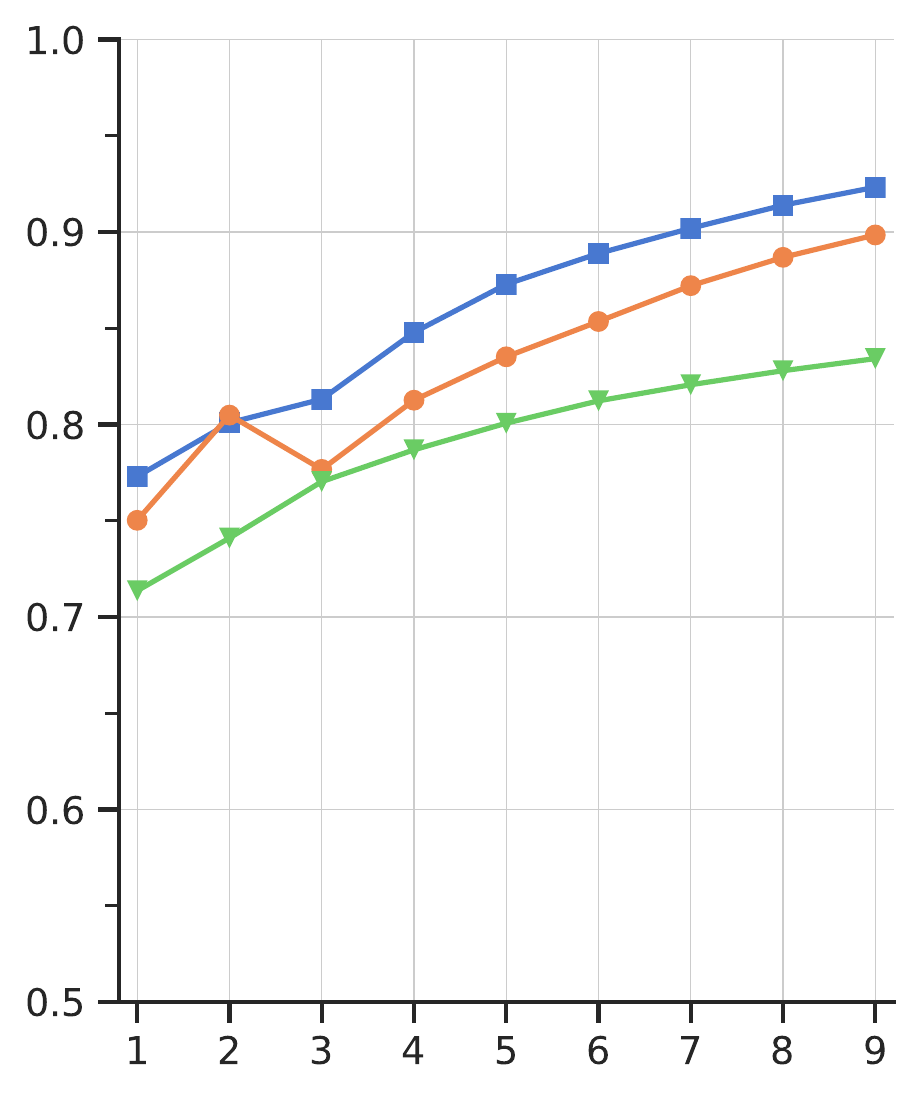}
        \caption{Reservoir~\cite{vitter1985random}}
        \label{fig:fd-dlib}
    \end{subfigure}
    \hfill
            \centering
    \begin{subfigure}[b]{0.24\linewidth}
        \centering
        \includegraphics[width=\linewidth]{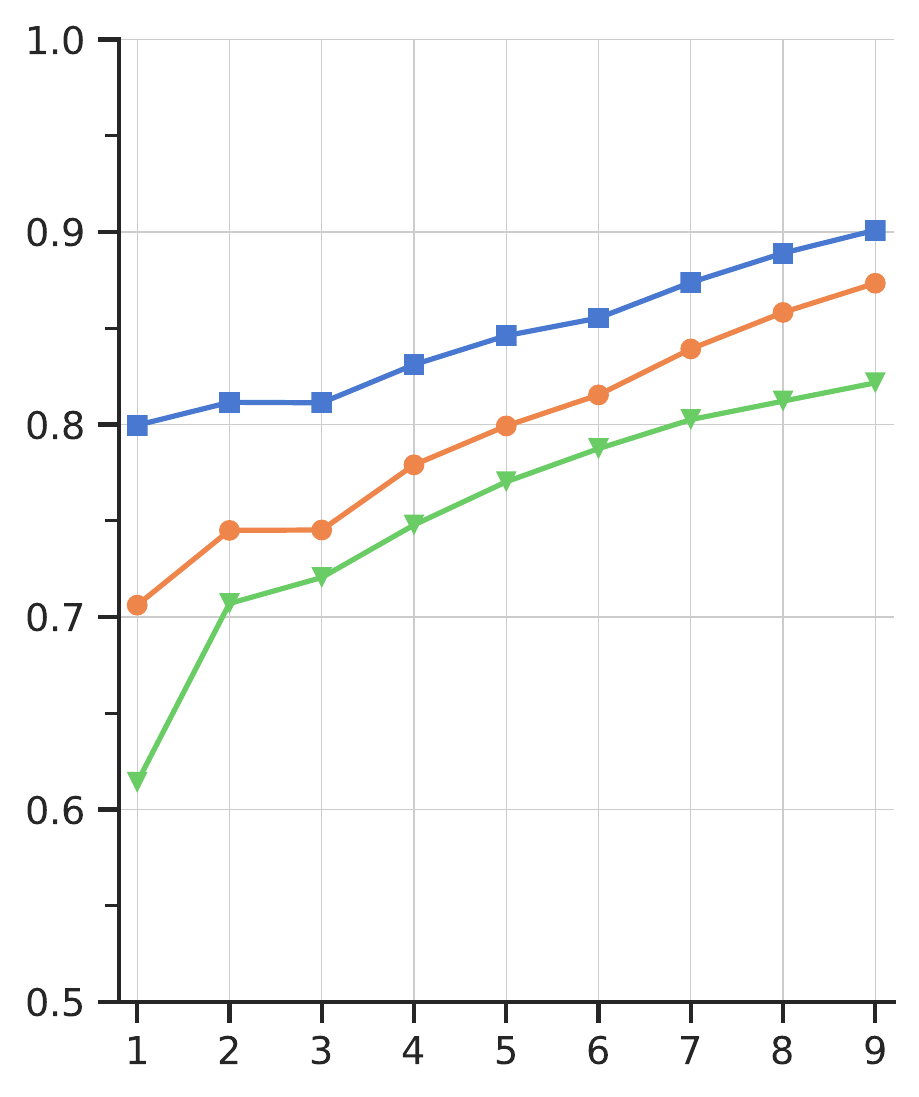}
        \caption{Random~\cite{chrysakis2020online}}
        \label{fig:fd-dlib}
    \end{subfigure}
    \hfill
\caption{Results obtained on the OpenLoris~\cite{she2019openlorisobject} dataset.}
\label{fig:results_openloris}
\end{figure*}

\begin{figure*}[th!]
    \centering
    \begin{subfigure}[b]{0.24\linewidth}
        \centering
        \includegraphics[width=\linewidth]{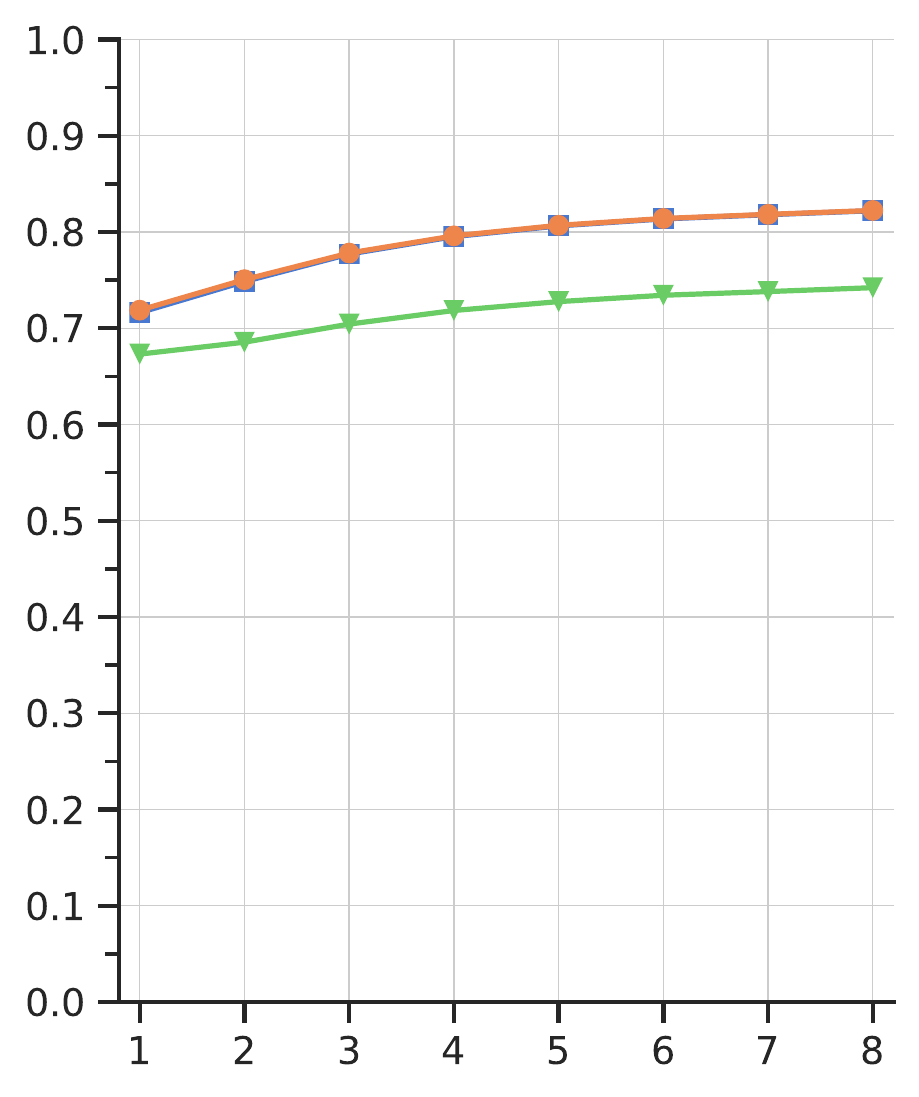}
        \caption{SLDA~\cite{hayes2020lifelong}}
        \label{fig:fd-dlib}
    \end{subfigure}
    \hfill  
    \centering
    \begin{subfigure}[b]{0.24\linewidth}
        \centering
        \includegraphics[width=\linewidth]{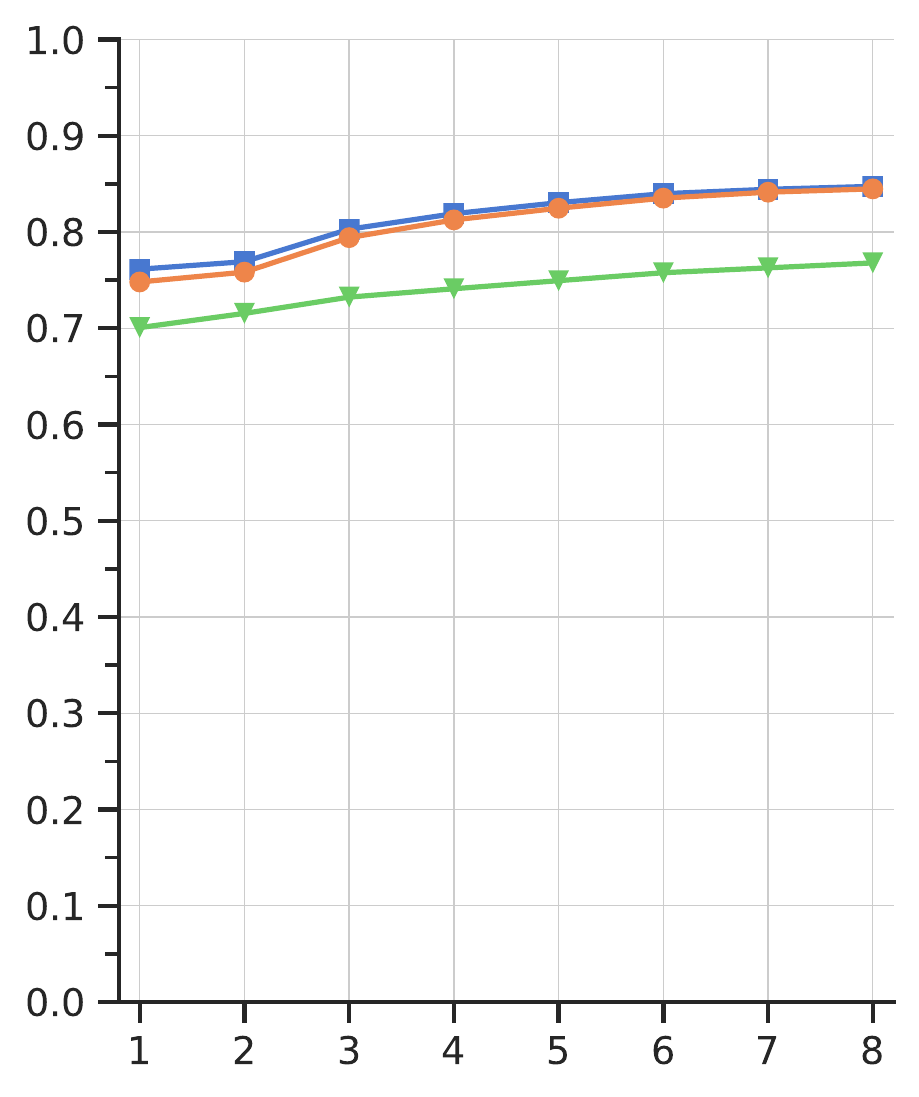}
        \caption{ExStream~\cite{hayes2019memory}}
        \label{fig:fd-dlib}
    \end{subfigure}
    \hfill   
        \centering
    \begin{subfigure}[b]{0.24\linewidth}
        \centering
        \includegraphics[width=\linewidth]{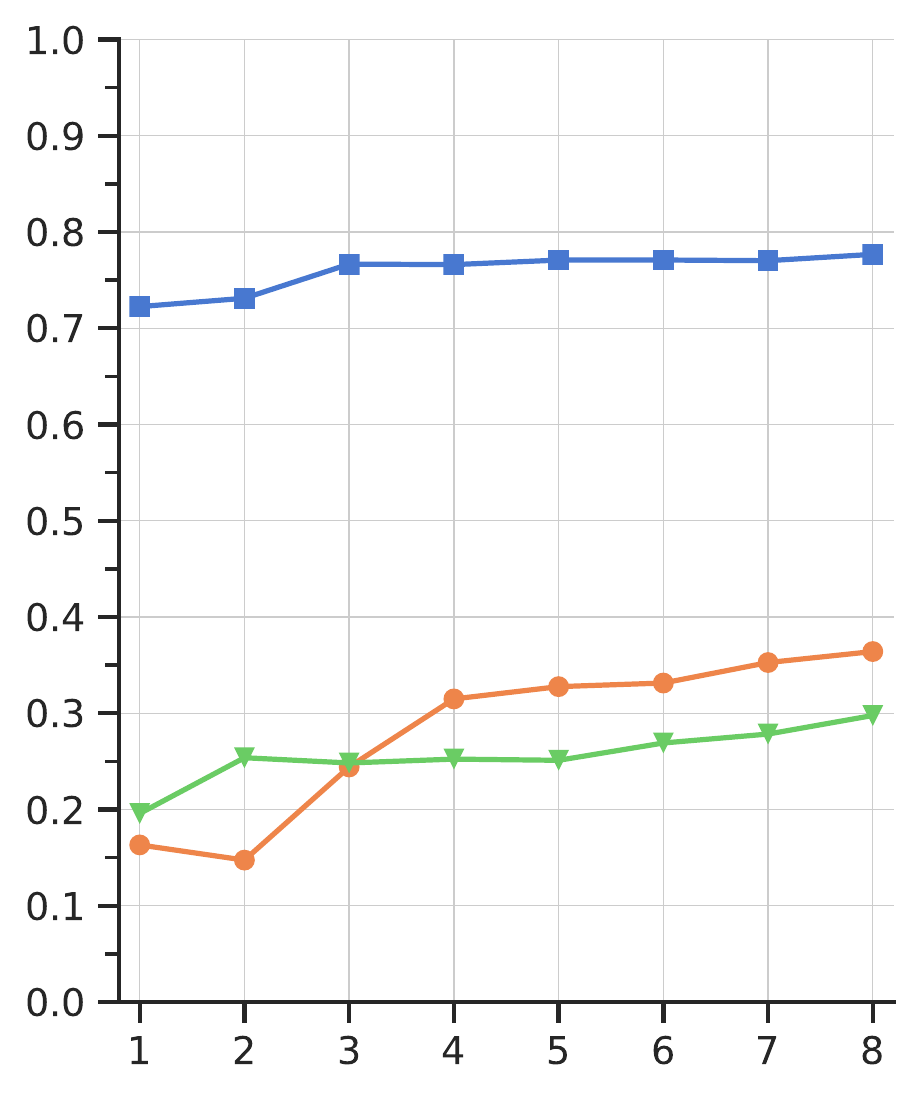}
        \caption{LwF~\cite{li2017learning}}
        \label{fig:fd-dlib}
    \end{subfigure}
    \hfill
        \centering
    \begin{subfigure}[b]{0.24\linewidth}
        \centering
        \includegraphics[width=\linewidth]{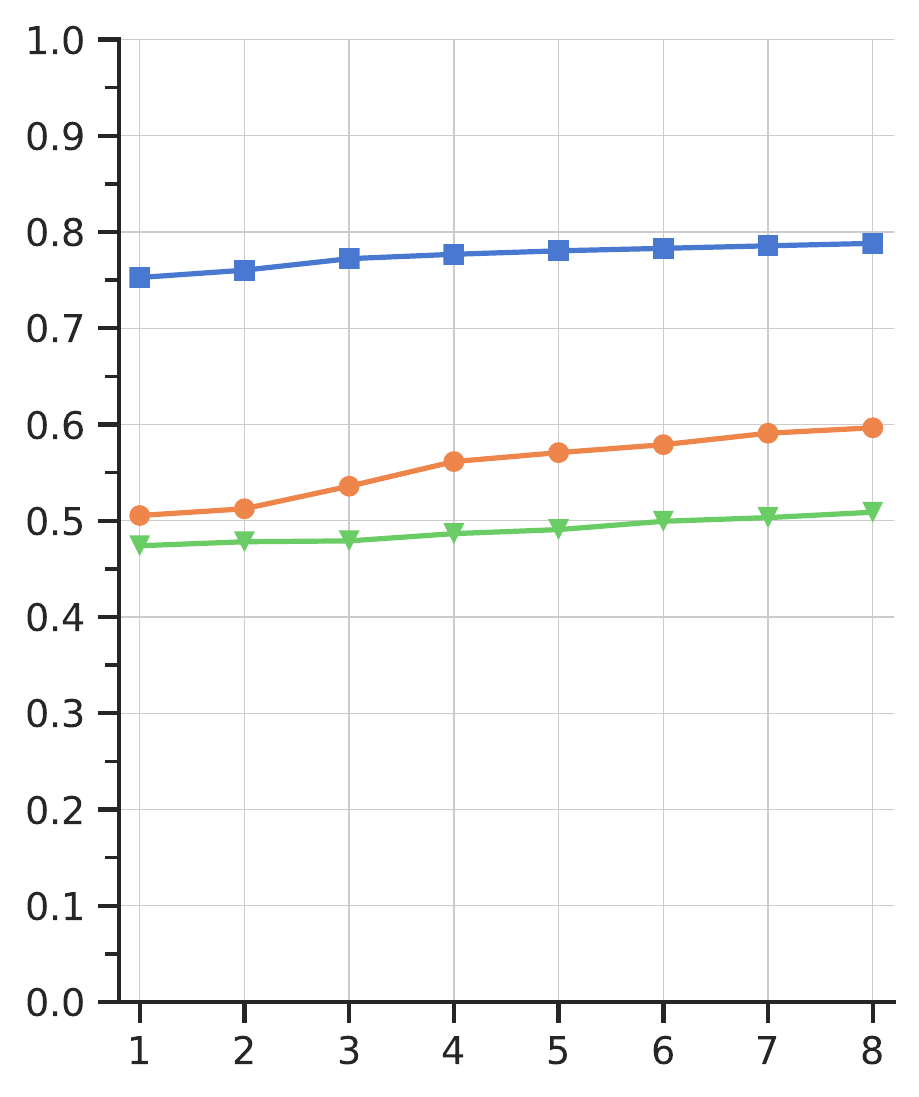}
        \caption{SSLAD~\cite{graffieti2022continual}}
        \label{fig:fd-dlib}
    \end{subfigure}
    \hfill
        \centering
    \begin{subfigure}[b]{0.24\linewidth}
        \centering
        \includegraphics[width=\linewidth]{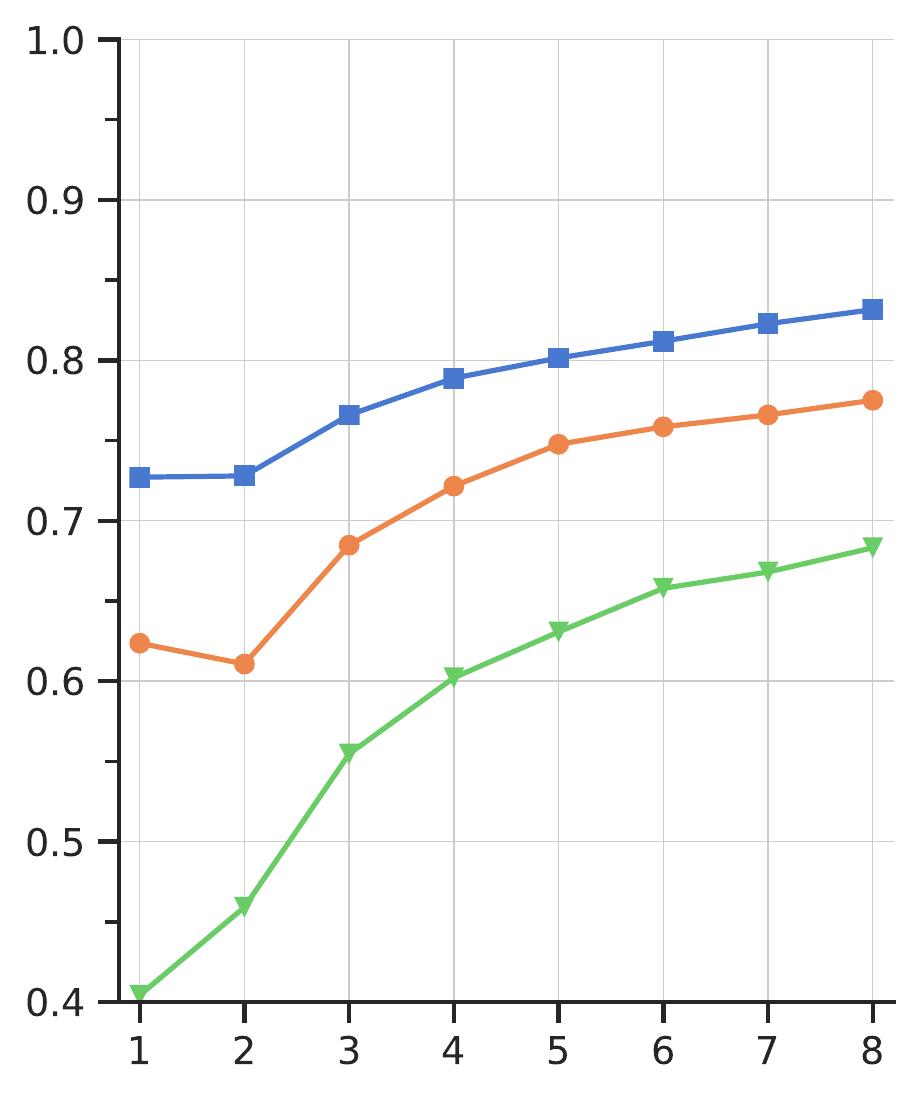}
        \caption{CBRS~\cite{chrysakis2020online} }
        \label{fig:fd-dlib}
    \end{subfigure}
    \hfill
        \centering
    \begin{subfigure}[b]{0.24\linewidth}
        \centering
        \includegraphics[width=\linewidth]{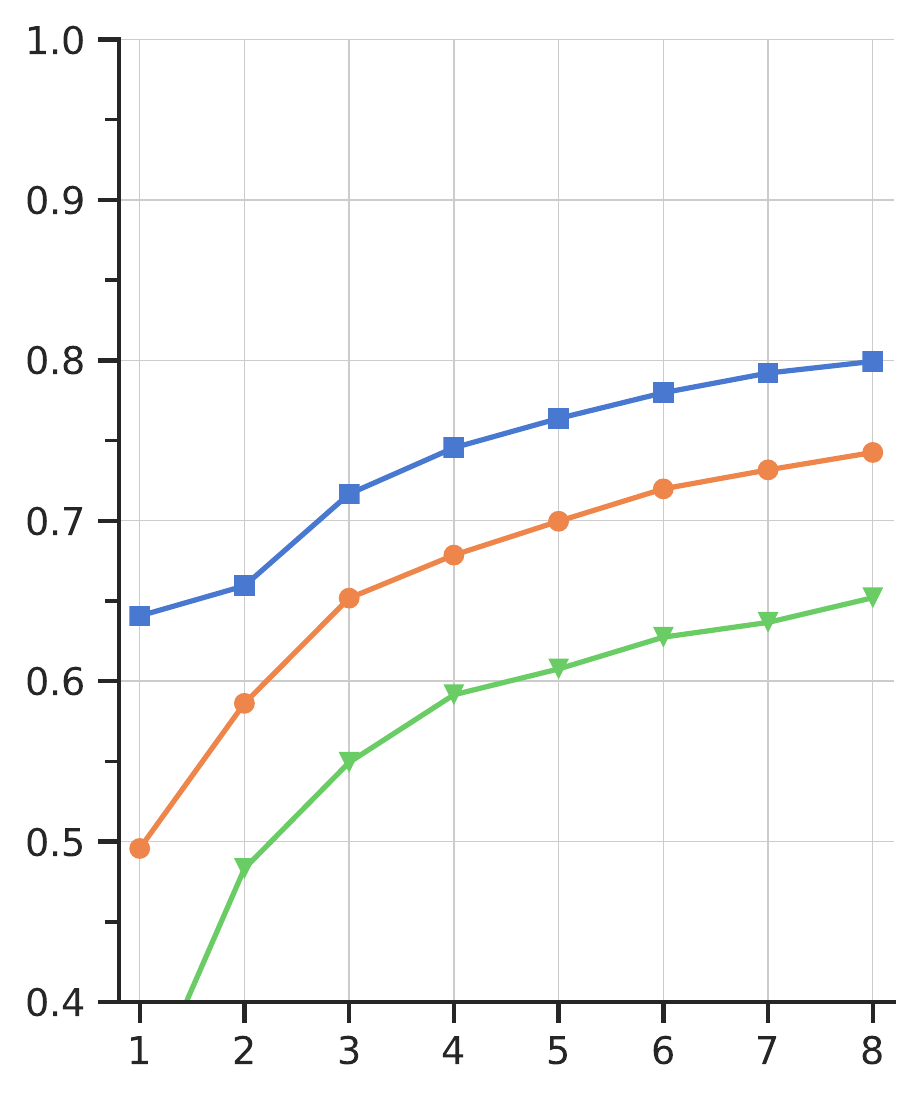}
        \caption{GSS~\cite{aljundi2019gradient}}
        \label{fig:fd-dlib}
    \end{subfigure}
    \hfill
        \centering
    \begin{subfigure}[b]{0.24\linewidth}
        \centering
        \includegraphics[width=\linewidth]{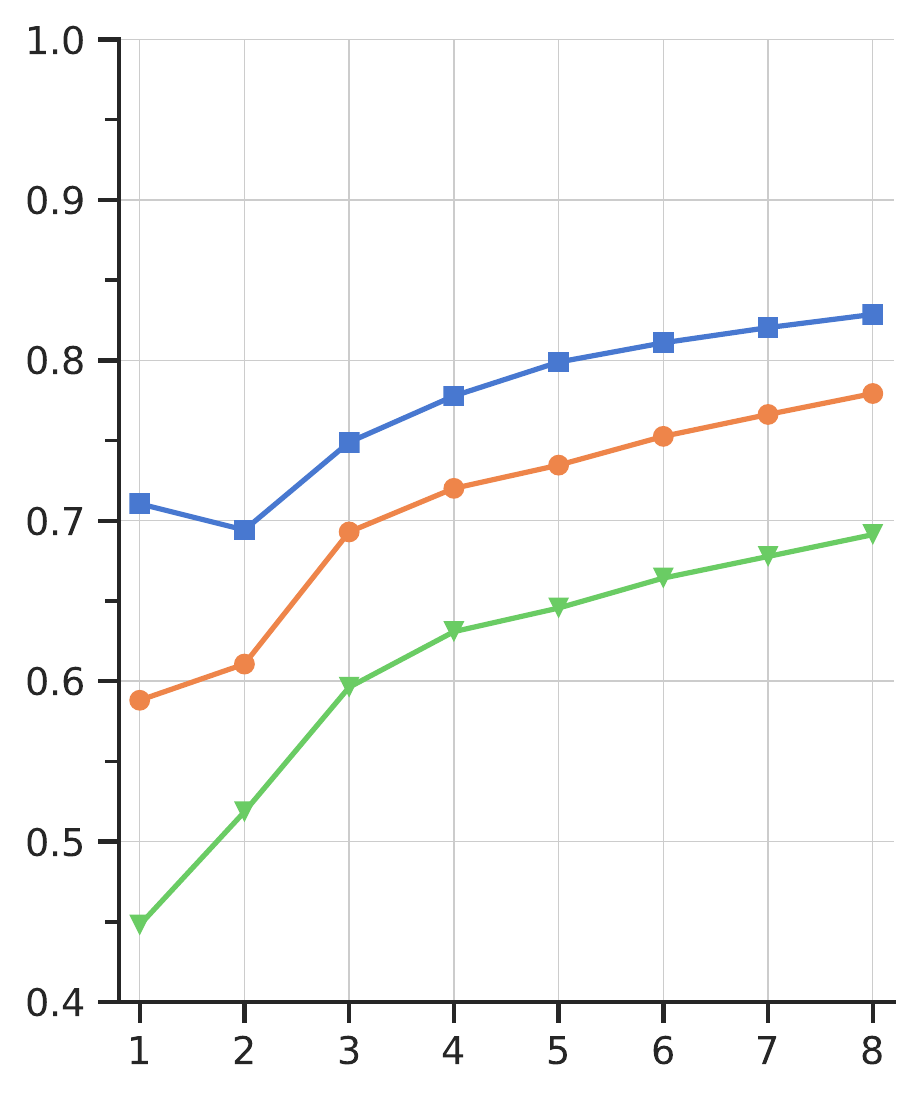}
        \caption{Reservoir~\cite{vitter1985random}}
        \label{fig:fd-dlib}
    \end{subfigure}
    \hfill
            \centering
    \begin{subfigure}[b]{0.24\linewidth}
        \centering
        \includegraphics[width=\linewidth]{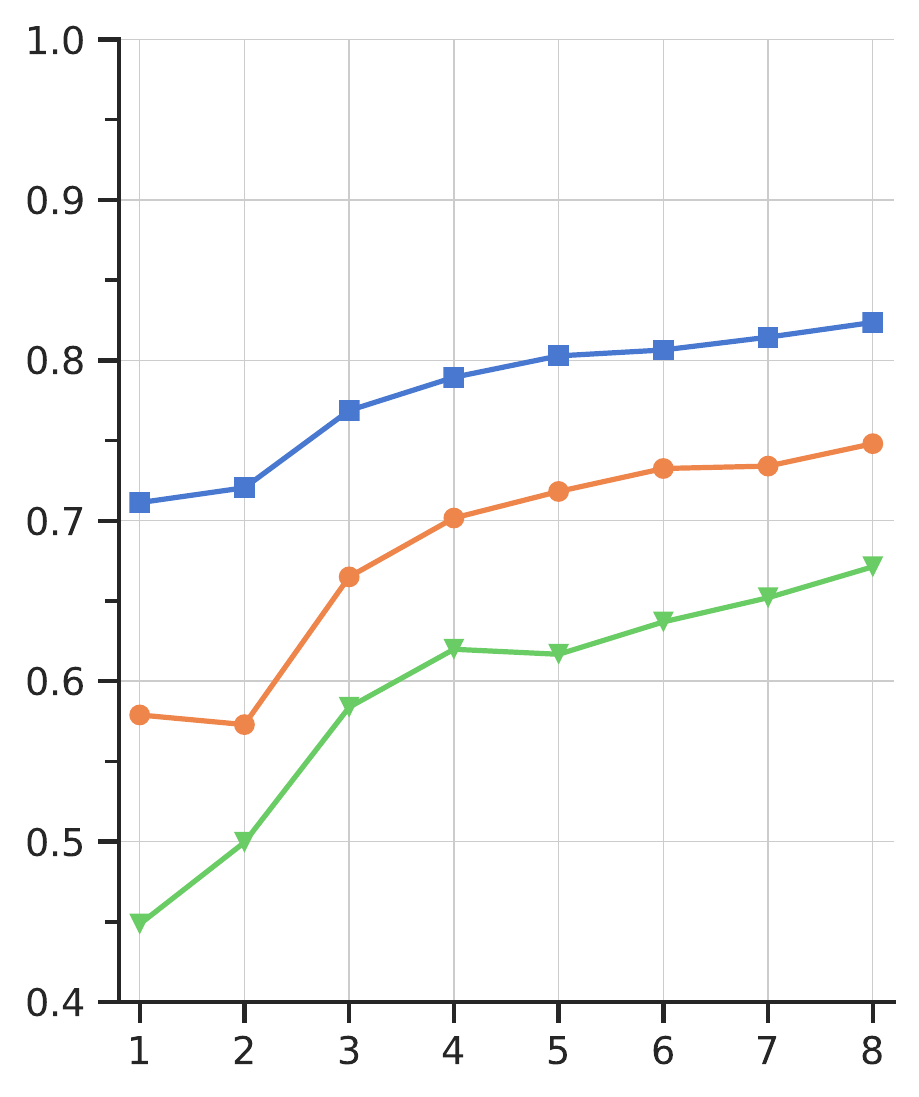}
        \caption{Random~\cite{chrysakis2020online}}
        \label{fig:fd-dlib}
    \end{subfigure}
    \hfill
\caption{Results obtained on the Core50~\cite{lomonaco2017core50} dataset.}
\label{fig:results_core}
\end{figure*}

\subsection{Metrics}
To assess the performance of the tested solutions, we rely on a metric used in similar recent works~\cite{graffieti2022continual}: the Average Mean Class Accuracy (AMCA) defined as:
\begin{equation}
    AMCA = \frac{1}{N_c \cdot N_{me}} \sum_{c = 1}^{N_{c}} \sum_{e = 1}^{N_{me}} \frac{corr_e^c}{tot_e^c}
\end{equation}
where the number of classes and macro-experience is $N_c$ and $N_{me}$, respectively. 
$corr_e^c$ and $tot_e^c$ are the number of correctly classified test images of class $c$ on macro-experience $e$, and the total number of test images of class $c$ in macro-experience $e$.
AMCA is useful to investigate the capability of a learning agent to incrementally learn, \textit{i.e.} to improve the classification ability over time, across different macro-experiences.

\subsection{Investigated Methods}
Many of the investigated methods are based on a combination of Deep Learning architectures and approaches that belong to different research fields, \textit{i.e.} Streaming~\cite{hoens2012learning}, Online~\cite{perry2011online} and Continual Learning~\cite{parisi2019continual}.

For each investigated method, we report a brief description in the following and, in the experimental evaluation, we use the implementation of the authors if available and the training parameters reported in the original paper.

ExStream~\cite{hayes2019memory} method, introduced in $2019$, essentially consists of a strategy to cluster buffers of prototypes of each class. Then, the last fully connected layer of a CNN is trained using all the samples stored in the buffers, following a batch-based learning procedure similar to the common Machine Learning paradigm. We observe that ExStream tends to extend the training time as buffers are filled since the model can do several iterations (epochs) on buffered data. The use of different buffers is capable of partially mitigating the problem of unbalanced data, especially when buffers are filled, but contrasts with the single-epoch learning paradigm (see Section~\ref{sec:introduction}).

In~\cite{hayes2020lifelong}, the authors revisit the streaming linear discriminant analysis, widely used in the data mining field, to handle streaming input data, with a method here referred to as SLDA. In particular, the authors propose to compute for each class a mean vector and a covariance matrix that is updated or used fixed after the base initialization. The covariance matrix is then used to compute the final prediction. Features are extracted with a pre-trained and frozen feature extractor, \textit{i.e.} ResNet-18~\cite{he2016deep}. We observe that the covariance matrix is sensitive to the class order, but it is less sensitive to catastrophic forgetting since its running means are independent for each class. Moreover, this method is not based on the common backpropagation-based training procedure, avoiding issues related to the batch size, their composition and to the choice of hyperparameters (optimizer, learning rate and so on).

Differently from the previous works, in~\cite{chrysakis2020online} authors investigate different strategies to populate the replay memory in the Continual Learning learning paradigm, without any assumptions about the input data stream (in particular, no i.i.d. assumptions and task boundaries), to train a conventional neural network architecture (ResNet-50~\cite{he2016deep}). In particular, the authors propose to select samples that are representative of all previously seen data for the replay buffer without the knowledge about task boundaries that correspond, in our benchmarks, to the macro-experiences. 
Authors investigate four different memory population strategies that we also consider in our experimental evaluation: 
i) \textit{Class-Balancing Reservoir Sampling} (CBRS): this strategy aims to preserve the distribution of each class of the training dataset, altering the input stream distribution so that unbalanced data distribution is mitigated. Specifically, as long as the memory is not filled, all the input instances are stored, otherwise, instances are stored relying on the status of the specific incoming class (largest, full and filled); 
ii) \textit{Gradient-based Sample Selection}~\cite{aljundi2019gradient} (GSS): this method greedily maximizes the variance of the samples contained in the replay memory, analyzing the gradient directions. A similarity analysis is conducted between the input instance and some randomly sampled stored instances. We observe the main drawback of this approach is the high computational load (depending on the GPU used) since several feed-forward operations are required for each new input and related comparisons.
iii) \textit{Reservoir}~\cite{vitter1985random}: the memory population strategy is divided into two steps: in the first, similar to the CBRS approach, all input instances are stored until the memory is filled; in the second one, the current input instance is stored with a probability of $\frac{m}{n}$, with $m$ as memory size and $n$ number of instances encountered so far.
iv) \textit{Random}: the current input instance is stored with $50$\% of probability in a randomly picked store location.

In our investigation, we test also the performance of the well-known Learning without Forgetting~\cite{li2017learning} (LwF) method, a Continual Learning algorithm, expressively designed to contrast the catastrophic forgetting, but without taking into consideration the aforementioned features of Natural Data Streams. In particular, LwF proposes the use of the knowledge distillation~\cite{gou2021knowledge}, consisting of two models, one of the current experience and one of the past one: the old model is frozen and used to distil knowledge into the current model.

Finally, we include in our analysis also on the work proposed in~\cite{graffieti2022continual}, developed for the SSLAD competition~\cite{sslad} (and then referred as SSLAD), developed for the Streaming Learning, that uses the CWR~\cite{maltoni2019continuous} algorithm to contrast the catastrophic forgetting~\cite{mccloskey1989catastrophic} and a double-memory to tackle the unbalanced data distribution in input, exploiting a reservoir mechanism in the first level of the memory. Unfortunately, the proposed method relies on the knowledge of the macro-experience boundaries, contrasting the assumptions about Natural Data Streams.

A similar investigation to our work is reported in~\cite{hayes2022online}: this analysis is focused on embedded devices, with limited memory and computational capacity. Then, experiments are based on efficient neural networks, such as EfficientNet~\cite{tan2019efficientnet} and MobileNet~\cite{sandler2018mobilenetv2}, but also common networks like ResNet-18~\cite{he2016deep}, and three different datasets: OpenLoris~\cite{she2019openlorisobject}, Places-365~\cite{lopez2020semantic} and Places-Long-Tail~\cite{liu2019large}. It is worth noting that input data are not organized in Natural Data Streams, and then the combined impact of temporal similarity and unbalanced data distribution is not evaluated.

\subsection{Experimental Results}
In all experiments, the experience size, \textit{i.e.} the amount of data that can be accumulated before triggering the learning process, is set to $10$, as also used in~\cite{chrysakis2020online,hayes2019memory}. When possible, we force learning agents to iterate on training data for only one epoch. All methods receive input images with a $128 \times 128$ spatial resolution. 
Finally, it is important to note that macro-experience boundaries are not provided during the training, while they are used only to compute the AMCA metric. 

Results obtained on the OpenLoris and Core50 datasets are reported in Figure~\ref{fig:results_openloris} and Figure~\ref{fig:results_core}, respectively.

In both, blue lines show the performance of each method with the balanced version of the two datasets as input. In this case, the whole set of training data is shuffled so the stream of input data is not temporally continuous, \textit{i.e.} is not a video stream but a stream of images without any temporal coherence. 
This setting is close to the common Machine Learning scenario, in which usually (but not mandatory) input data is balanced across available classes and the shuffling operation is performed before each epoch~\cite{chicco2017ten}.
As expected, in this scenario a large part of the investigated methods achieve the best accuracy across other evaluations: the models continuously learn across different macro-experiences achieving a high final AMCA ($~90\%$ with OpenLoris and $~80\%$ with Core50), with an improvement with respect to the first macro-experience of about $~15\%$ with both datasets.  

The orange line represents the performance of the evaluated strategies receiving as input the balanced version of the datasets, but without any shuffling operation: this aims to highlight the impact of the time similarity.
From a general point of view, all the models obtain worse results w.r.t. the previous experiment, with a worsening of about $~10\%$ on OpenLoris and $~20\%$ on Core50. SLDA and ExStream represent an exception since we observe that their performance is not affected by temporal similarity. In particular, SLDA is based on a pre-trained and blocked feature extractor and only a covariance matrix is updated, limiting the overfitting phenomena that can occur instead during the usual training of deep learning architectures, in which all the weights of the model are updated. Besides, we observe that in ExStream the model is fine-tuned using the instances stored in several replay buffers (one for each class), limited in size, shuffled before each iteration.
Moreover, we note that LwF strongly suffers temporal similarity in input data, confirming that a Continual Learning model without any mechanism (such as but not only a replay memory) to contrast the challenges introduced by Natural Data Streams, performs poorly in this specific scenario, with a worsening of about $-70\%$ on OpenLoris and about $-40\%$ on Core50.

Finally, the green line shows the behaviour of the investigated methods in presence of unbalanced data distributions and temporal similarity, \textit{i.e.} Natural Data Stream. 
On Core50 dataset, we observe that the performance of all the methods tested degraded significantly, including SLDA and ExStream, while on OpenLoris the performances remain almost unchanged, probably due to the fact that in OpenLoris dataset the test frames are similar to the training ones (they are sampled from the same video sequence), and then the classification task is easier. This consideration is also supported by the greater performance generally achieved on OpenLoris dataset.
It is also worth noting that even CBRS, which is specifically designed to contrast data imbalance, performs way poorer in this setting than in the previous ones, confirming the difficulties posed by Natural Data Streams.


\begin{figure}
    \centering
    \includegraphics[width=1\linewidth]{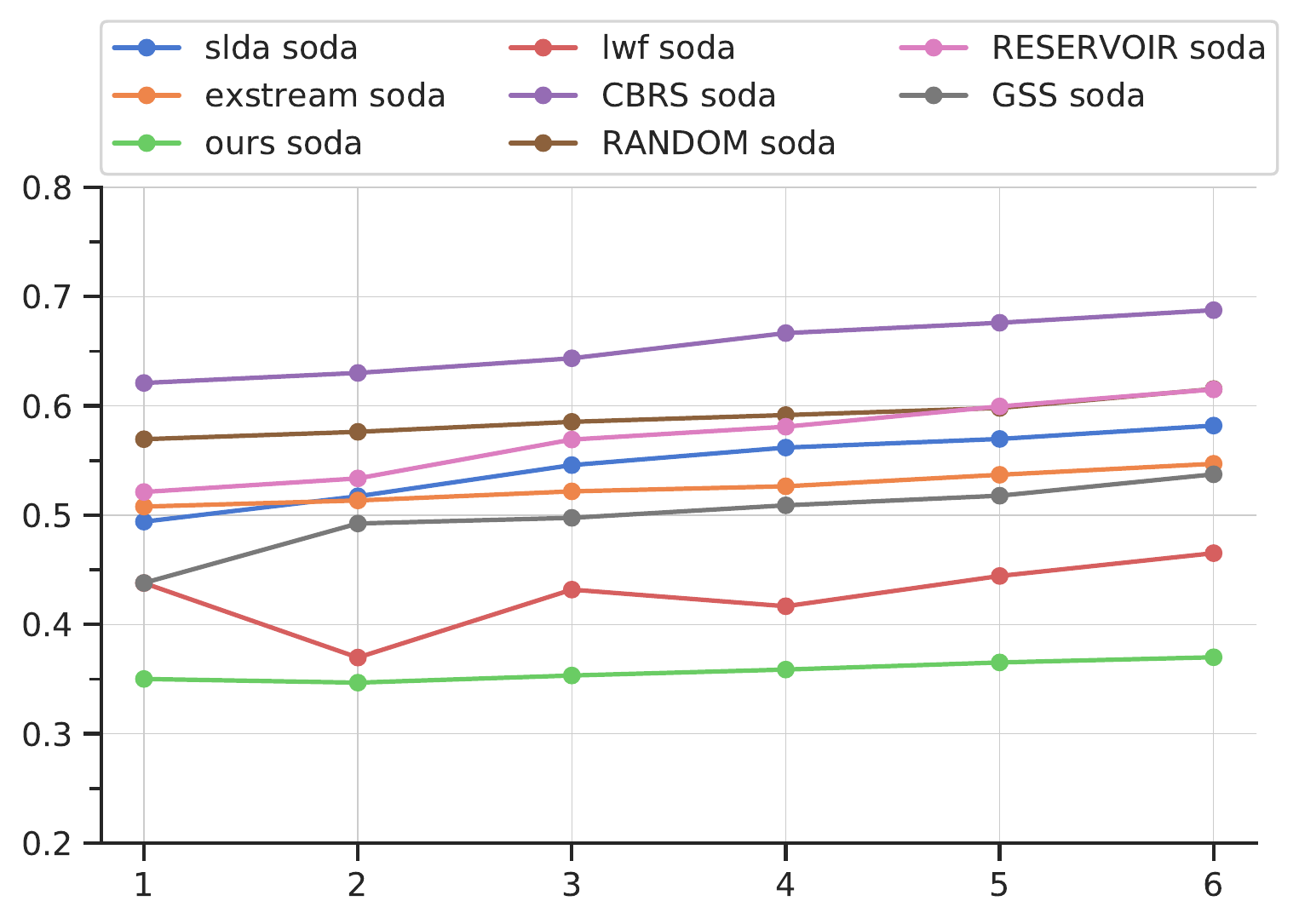}
    \caption{Results obtained on the Soda10~\cite{han2021soda10m} dataset.}
    \label{fig:soda_results}
\end{figure}

Finally, we run the methods on the Soda10 dataset, collecting the results reported in Figure~\ref{fig:soda_results}. Probably Soda10 represent the most challenging test set, since it combines difficulties for the classification task related to a real-world acquisition (\eg low quality images, extreme object variability, persistent occlusions) with the challenges introduced by Natural Data Streams.
Indeed, as shown, tested methods are able to incrementally learn with only a limited margin, confirming the challenges introduced by Natural Data Streams. Secondly, CBRS method overcomes the SLDA approach, probably due to the presence of a unblocked neural network and replay memory. 


\section{Conclusion}
In this paper, several methods available in the literature have been tested on Natural Data Stream, \ie streams of data mainly characterized by temporal similarity and unbalanced data distribution. Experimental results suggest that this kind of streams contain challenges to be addressed in future research works, in order to improve the performance of AI agents operating in real-world scenarios. 

\ifCLASSOPTIONcaptionsoff
  \newpage
\fi

\bibliographystyle{IEEEtran}
\bibliography{root_journal}

\end{document}